\documentclass[letterpaper]{article} 
\usepackage{aaai2026}  
\usepackage{times}  
\usepackage{helvet}  
\usepackage{courier}  
\usepackage[hyphens]{url}  
\usepackage{graphicx} 
\usepackage{natbib}  
\usepackage{caption} 
\usepackage{algorithm}
\usepackage{algorithmic}

\usepackage{newfloat}
\usepackage{listings}

\usepackage{amsmath,amssymb,amsfonts}
\usepackage[english]{babel}
\usepackage{algorithmic}
\usepackage{graphicx}
\usepackage{textcomp}
\usepackage{booktabs}
\usepackage{lineno}
\usepackage{multirow}
\usepackage{xcolor}
\usepackage{amsmath, amssymb}
\usepackage{graphicx}
\usepackage{booktabs}
\usepackage{algorithm}
\usepackage{algorithmic}
\usepackage{caption}

\usepackage{amsmath}
\usepackage{listings}
\usepackage[dvipsnames]{xcolor}
\usepackage{enumitem}
\usepackage{booktabs}
\usepackage{colortbl}
\usepackage{makecell}
\usepackage{graphicx}
\usepackage{subfig}
\usepackage{subcaption}
\usepackage{color}

\DeclareCaptionStyle{ruled}{labelfont=normalfont,labelsep=colon,strut=off} 
\floatstyle{ruled}
\newfloat{listing}{tb}{lst}{}
\floatname{listing}{Listing}

\newcolumntype{R}{@{\extracolsep{0.17cm}}c@{\extracolsep{0pt}}}%

\newcommand{\specialcellbold}[2][R]{%
  \bfseries
  \begin{tabular}[#1]{@{}c@{}}#2\end{tabular}%
}
\newcommand{\specialcellnormal}[2][R]{%
  \begin{tabular}[#1]{@{}c@{}}#2\end{tabular}%
}

\title{Fusing Biomechanical and Spatio-Temporal Features for Fall Prediction: Characterizing and Mitigating the Simulation-to-Reality Gap}
\author {
    Md Fokhrul Islam\textsuperscript{\rm 1},
    Sajeda Al-Hammouri\textsuperscript{\rm 2},
    Christopher J. Arellano\textsuperscript{\rm 2, 3},
    Kavan Hazeli\textsuperscript{\rm 2, 4},
    Heman Shakeri\textsuperscript{\rm 1}
}
\affiliations {
    \textsuperscript{\rm 1}School of Data Science, University of Virginia\\
    \textsuperscript{\rm 2}Biomedical Engineering, University of Arizona\\
    \textsuperscript{\rm 3}Department of Orthopaedic Surgery, University of Arizona\\
    \textsuperscript{\rm 4}Aerospace and Mechanical Engineering, University of Arizona\\
    \{sdh4py, hs9hd\}@virginia.edu

    }

\usepackage{bibentry}

\begin{document}

\maketitle

\begin{abstract}
Falls are a leading cause of injury and loss of independence among older adults. Vision-based fall prediction systems offer a non-invasive solution to anticipate falls seconds before impact, but their development is hindered by the scarcity of available fall data. Contributing to these efforts, this study proposes the Biomechanical Spatio-Temporal Graph Convolutional Network (BioST-GCN), a dual-stream model that combines both pose and biomechanical information using a cross-attention fusion mechanism. Our model outperforms the vanilla ST-GCN baseline by 5.32\% and 2.91\% F1-score on the simulated MCF-UA stunt-actor and MUVIM datasets, respectively. The spatio-temporal attention mechanisms in the ST-GCN stream also provide interpretability by identifying critical joints and temporal phases.
However, a critical simulation-reality gap persists. While our model achieves an 89.0\% F1-score with full supervision on simulated data, zero-shot generalization to unseen subjects drops to 35.9\%. This performance decline is likely due to biases in simulated data, such as `intent-to-fall' cues. For older adults, particularly those with diabetes or frailty, this gap is exacerbated by their unique kinematic profiles. To address this, we propose personalization strategies and advocate for privacy-preserving data pipelines to enable real-world validation. Our findings underscore the urgent need to bridge the gap between simulated and real-world data to develop effective fall prediction systems for vulnerable elderly populations.

\end{abstract}

\section{Introduction}
Falls represent a significant public health challenge for older adults, contributing substantially to morbidity, mortality, and healthcare costs \cite{who2021falls}. Annually, 28–35\% of individuals over 65 experience a fall, making it the second leading cause of unintentional injury death worldwide and the primary cause for this demographic \cite{campbell2005randomised}. In the United States alone, over a quarter of older adults fall yearly, creating an economic burden exceeding \$50 billion annually \cite{adam2024impact, peterson2021average}. This burden is concentrated in high-risk groups like older adults with diabetes, who have a 39\% annual fall incidence \cite{vinik2017aging}. With the global population over 65 projected to reach 1.5 billion by 2050, the impact of falls is expected to grow dramatically.
Current fall management technologies are divided into two paradigms: post-fall detection and long-term risk assessment. Detection systems use wearable (e.g., accelerometers) or ambient (e.g., cameras) sensors to alert caregivers after a fall has occurred, making them inherently reactive \cite{ren2019research, aziz2017, mastorakis2018}. Long-term risk assessment uses clinical tools like the Timed Up and Go (TUG) test to identify at-risk individuals over months, informing preventative strategies like exercise programs \cite{ren2019research, campbell2005randomised, chen2020}. However, these approaches cannot predict a fall seconds before impact. This reveals a critical need for systems that can predict an imminent fall, which could trigger immediate interventions (e.g., wearable airbags) to mitigate injury \cite{ren2019research, mackay2021fear}.
The feasibility of imminent prediction rests on the existence of measurable kinematic precursors. Falls are processes preceded by detectable instability, with distinct differences in kinematics and neuromuscular patterns between fallers and non-fallers \cite{robinovitch2013biomechanics, ting2007neuromechanics}. Analysis of real-world falls shows an average of 583 milliseconds from the loss of balance to impact, a narrow window for intervention \cite{robinovitch2013biomechanics}. Computer vision, combined with markerless Human Pose Estimation (HPE), offers a non-invasive method to continuously capture the full-body kinematics needed for prediction \cite{ren2019research, wang2019}. HPE algorithms extract 3D joint locations from video, creating skeletal time-series data from which biomechanical features like center of mass (COM) velocity and joint angles can be derived \cite{bazarevsky2020blazepose}.
Deep learning models are well-suited for analyzing these complex spatio-temporal pose sequences \cite{wang2019}. Architectures like LSTMs, TCNs, and Spatio-Temporal Graph Convolutional Networks (ST-GCNs) have shown promise \cite{hochreiter1997long, alhammouri2025fall, shi2015, yan2018spatial}. ST-GCNs are particularly powerful as they model the human skeleton as a graph, simultaneously learning spatial and temporal dependencies \cite{yan2018spatial}. However, a significant bottleneck is the scarcity of real-world fall data from older adults \cite{klenk2016opportunities}. This necessitates reliance on simulated fall datasets, typically using young actors, which introduces critical simulation bias. Simulated falls contain `intent-to-fall' cues and involve different biomechanics than genuine falls in frail older adults, leading to distinct kinematic patterns that do not generalize well \cite{auvinet2010, robinovitch2013biomechanics, klenk2016simulated}. Models trained on this biased data often exhibit poor generalization, a `simulation-reality gap' that hinders practical application and clinical trust~\cite{wang2018deep}.

The primary objective of this work is to develop and evaluate an interpretable, dual-stream deep learning model for imminent fall prediction using simulated datasets. We aim to quantitatively characterize the `simulation-reality gap'—the performance drop on unseen simulated subjects (zero-shot generalization)—and investigate few-shot personalization as a strategy to adapt the model. We propose a dual-stream architecture fusing an ST-GCN stream for raw pose data with a BiLSTM stream for engineered biomechanical features. Integrated spatio-temporal attention mechanisms in the ST-GCN stream will enhance interpretability by highlighting influential joints and temporal segments. We hypothesize that the zero-shot generalization gap stems from models learning simulation-specific artifacts (`intent-to-fall' cues) and that few-shot fine-tuning can significantly improve performance on new subjects, offering a potential deployment pathway.

The main contributions of this paper are as follows:
\begin{itemize}
    \item Proposing and evaluating an enhanced, interpretable dual-stream deep learning model, integrating ST-GCNs with attention mechanisms for pose sequence analysis and Bidirectional LSTMs (BiLSTMs) for biomechanical feature modeling, aimed at improving vision-based imminent fall prediction.
    \item Demonstrating the model's significant performance improvements over baseline approaches on benchmark simulated fall datasets, while quantitatively characterizing the substantial gap between zero-shot generalization and full supervision performance, and thereby underscoring the critical role of personalization strategies in bridging the simulation-reality gap for real-world application.
    \item Providing empirical support for the hypothesis that this generalization gap is significantly driven by simulation bias present in common simulated datasets.
\end{itemize}


\section{Background and Related Work}
\label{sec:literature_review}

~\\\noindent\textbf{Sensing Modalities and Pose Estimation}
Fall prediction research utilizes various sensor modalities, each with distinct trade-offs. Wearable sensors (e.g., accelerometers) offer high accuracy but suffer from user compliance issues~\cite{aziz2017}. Ambient systems (e.g., pressure mats) are non-intrusive but have limited predictive power~\cite{rougier2011}. Vision-based approaches provide a rich, non-invasive data source but raise privacy concerns~\cite{mastorakis2018}. Within the vision-based domain, pose estimation is a cornerstone for analyzing human movement. While alternatives like OpenPose~\cite{cao2019openpose} and HRNet~\cite{sun2019hrnet} offer higher accuracy at a greater computational cost, we employ BlazePose~\cite{bazarevsky2020blazepose} for its balance of real-time performance and precision (86.9\% mAP on COCO). However, its performance can degrade with occlusions and atypical body proportions, challenges particularly relevant when monitoring elderly subjects.
~\\\noindent\textbf{Human Activity Recognition (HAR)}
Methodologies for HAR have evolved to better capture movement dynamics. Early approaches using Long Short-Term Memory (LSTM) networks modeled temporal dependencies but struggled to represent inter-joint spatial relationships explicitly~\cite{hochreiter1997, nunez2018}. To incorporate domain-specific knowledge, biomechanical feature engineering is critical. This involves extracting kinematic parameters (e.g., joint angles), stability metrics (e.g., center of mass), and inter-segment coordination patterns~\cite{thomas2022winter}. Research has shown that specific markers, such as trunk angular velocity~\cite{bourke2008} and center of mass stability~\cite{maki1994, pai2003}, are highly predictive of falls. Integrating these features enhances both model performance and clinical interpretability~\cite{palmerini2015}. More recently, graph-based models directly address the limitations of earlier methods by representing the human skeleton as a graph, where joints are nodes and limbs are edges. The seminal ST-GCN~\cite{yan2018spatial} established a paradigm for simultaneously learning spatial and temporal features. Subsequent advances, including multi-scale operations~\cite{liu2020disentangling} and adaptive graph learning~\cite{shi2019twobranch}, have further improved performance in modeling coordinated movements. Given their efficacy, we explore this paradigm in our proposed architecture.
~\\\noindent\textbf{Data Bias in Fall Research}
The validity of fall prediction models is heavily influenced by data biases that can compromise generalization. \textit{Selection bias} arises when study populations are not representative, such as over-representing healthy, younger subjects in fall research~\cite{young2019selection}. \textit{Measurement bias} can stem from instrumentation; for instance, camera positioning alone can alter fall detection accuracy by up to 23\%~\cite{lin2021measurement}. Furthermore, the \textit{Hawthorne effect}, where subjects modify their behavior because they are being observed, can significantly alter movement patterns, with participants exhibiting up to 17\% less postural sway during explicit monitoring~\cite{kumar2018hawthorne}.
A key challenge is \textit{intent bias}, where consciously performed actions differ kinematically from spontaneous ones. Simulated falls, especially by actors, exhibit non-representative preparatory movements and longer reaction times compared to unexpected falls~\cite{Casilari2022analytical, robinovitch2013biomechanics}. Consequently, models often learn these performance artifacts, which explains the significant performance gap between personalized and generalized models.

\section{Methodology}
\label{sec:method}

\subsection{Background and Problem Formulation}
\label{sec:background}

In automated fall management, a distinction is made between fall detection and fall prediction. Fall detection systems are reactive, identifying a fall event during or immediately after its occurrence. This task is formulated as a binary classification $f_d(S_t) \rightarrow \{0,1\}$, where $S_t$ is sensor data at time $t$. In contrast, fall prediction is a proactive approach that aims to forecast an impending fall, framed as a probabilistic estimation problem: $f_p(S_{t-w:t}) \rightarrow P(F_{t+h})$. Here, a sequence of observations $S_{t-w:t}$ over a history window $w$ is used to predict the probability of a fall $P(F_{t+h})$ within a future prediction horizon $h$, typically 0.5--2 seconds~\cite{ren2019research}.

This study addresses fall prediction using vision-based human action recognition. Pose data were extracted using pose model~\cite{bazarevsky2020blazepose}, which identifies $33$ anatomical landmarks per frame. The framework was configured for tracking stability (confidence threshold = $0.6$, model complexity = 2), yielding a pose representation of dimension $N \times 33 \times 3$ for a sequence of $N$ frames.
To establish ground-truth labels, we implemented a temporal offset strategy with prediction horizons at 2\,s, 1\,s, 0.5\,s, and 0\,s (fall onset) before fall occurred. For fall sequences, frames within these horizons were labeled positive (1); all frames in non-fall sequences were labeled negative (0). Continuous pose data was segmented using a 90-frame (3\,s) sliding window. For training, a stride of 15 frames (0.5\,s) was used for data augmentation, while non-overlapping windows were used during testing to ensure an independent evaluation.

\subsection{Feature Sets}
Our proposed dual-stream architecture processes two distinct input streams: raw skeletal pose coordinates and engineered biomechanical features derived from these poses. The first input stream is the raw pose data $\mathcal{P} \in \mathbb{R}^{90 \times 33 \times 3}$, extracted via MediaPipe Pose \cite{bazarevsky2020blazepose}, representing 90 frames of 33 anatomical landmarks in 3D space for each window. We apply a Savitzky--Golay filter (window length 7, polynomial order 3) to smooth the pose trajectories over time and normalize all coordinates by the subject’s height \cite{kuster2016} to account for inter-individual scale differences.
The second input stream consists of 45 biomechanical features $\mathbf{B} \in \mathbb{R}^{90 \times 45}$, derived from the raw pose data.  
The biomechanical features are as follows: \textit{27 angular positions} representing the three Euler angles for nine major body segments; \textit{9 centroid locations} detailing the 3D coordinates of the upper-body, lower-body, and whole-body center of mass; a \textit{single yaw trunk angle} for torso rotation; and \textit{8 features from 2D image coordinates} of key hip and shoulder landmarks. This feature set is grounded in established biomechanical principles\cite{alhammouri2025fall}, including a gravity-aligned anatomical frame and a mass distribution model based on Plagenhoef’s anthropometric data~\cite{plagenhoef1983}, to effectively capture indicators of balance and instability. 

\subsection{Proposed Model Architecture}
\begin{figure*}[t]
    \centering
    \includegraphics[width=\textwidth]{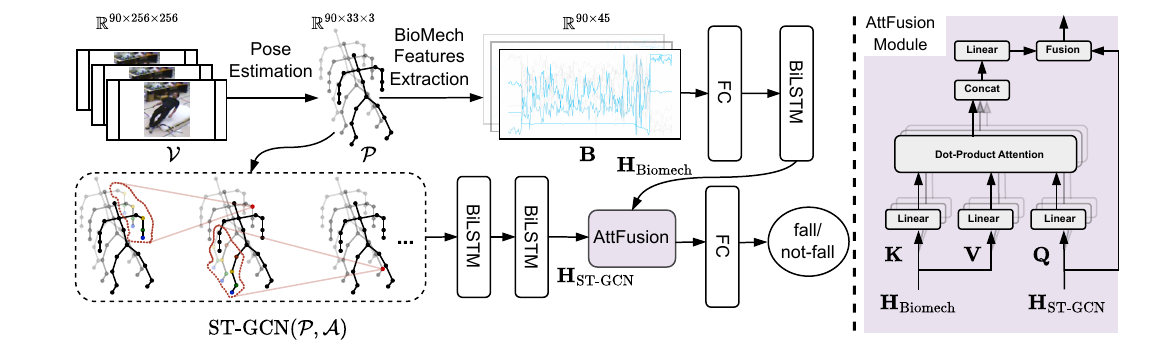}
    \caption{Fall Prediction Model Pipeline. The system extracts 3D pose landmarks from video, segments them, and feeds them into a dual-stream network. One stream uses ST-GCN for pose dynamics; the other uses a BiLSTM for engineered biomechanical features. An attention mechanism fuses features from both streams, followed by fully connected layers for fall probability prediction.
    }
    \label{fig:model_pipeline}
\end{figure*}

We propose the Biomechanical Spatio-Temporal Graph Convolutional Network (BioST-GCN), a dual-stream neural network architecture designed to synergistically process skeletal pose data and biomechanical features for effective fall prediction. This approach combines the representational power of graph convolutional networks with domain-specific biomechanical knowledge. To capture both spatial relationships and temporal dynamics critical for identifying fall precursors. The model is fed two distinct input streams per 90-frame window: a pose data tensor, $\mathcal{P}$ representing the 3D coordinates of 33 body landmarks, and a corresponding matrix of biomechanical features, $\mathbf{B}$.
Figure \ref{fig:model_pipeline} illustrates the comprehensive pipeline of our proposed model.

\subsubsection{ST-GCN Stream}
The ST-GCN stream processes raw pose data to capture both spatial relationships among the 33 keypoints and temporal dynamics across the 90 frames. We construct a spatial graph where each joint serves as a node, and edges represent the natural human skeletal structure (e.g., connections between joints). This graph is encoded in an adjacency matrix $\mathcal{A} \in \mathbb{R}^{33 \times 33}$, which defines the connectivity between joints. Following Yan et al. \cite{yan2018spatial}, the ST-GCN leverages this adjacency matrix to model joint interactions through graph convolutions.

The ST-GCN architecture consists of two blocks with channel configurations of 64 and 128, respectively, allowing the network to learn increasingly abstract features. Residual connections between blocks mitigate vanishing gradients, ensuring stable training in deeper layers. Each block applies spatio-temporal convolutions using the partition strategy from \cite{yan2018spatial}, which divides the graph into subsets (e.g., root, centripetal, centrifugal) to capture different motion patterns, such as inward and outward joint movements. These convolutions are followed by batch normalization and ReLU activation.

The output of the ST-GCN stream is a feature vector $\mathbf{H}_{\text{ST-GCN}} \in \mathbb{R}^{1 \times 80}$, obtained after global average pooling to aggregate spatial and temporal features. This vector encapsulates the spatio-temporal patterns relevant to fall prediction.
\begin{equation}
    \mathbf{H}_{\text{ST-GCN}} = \text{ST-GCN}(\mathcal{P}, \mathcal{A})
\end{equation}
where $\mathcal{P} \in \mathbb{R}^{90 \times 33 \times 3}$ is the input pose sequence, and $\mathcal{A}$ is the adjacency matrix.

\subsubsection{Body Attention Mechanism}
\label{sec:body_attention}
To dynamically weigh the biomechanical importance of body joints, we enhance the standard ST-GCN graph convolution with a learnable self-attention map. Classical ST-GCNs employ a static adjacency matrix $\mathcal{A}_k$; in contrast, we introduce a trainable attention matrix $\mathcal{M}_k \in \mathbb{R}^{33\times 33}$ for each input sequence. This attention map is produced by integrating a Convolutional LSTM (ConvLSTM) over the intermediate features of the ST-GCN blocks, capturing the spatio-temporal evolution of joint interactions. The effective adjacency at hop $k$ is computed as $\widetilde{\mathcal{A}}_k = \mathcal{A}_k \odot \mathcal{M}_k$. Consequently, the graph convolution operation at layer $\ell$ is expressed as:
\begin{equation}
\mathbf{F}^{(\ell+1)} = \sum_{k=1}^{K} \left(\mathbf{D}^{(\ell)}_k\right)^{-\frac{1}{2}} \widetilde{\mathcal{A}}^{(\ell)}_k \left(\mathbf{D}^{(\ell)}_k\right)^{-\frac{1}{2}} \mathbf{F}^{(\ell)} \mathbf{W}^{(\ell)}_k
\end{equation}
where $\mathbf{F}^{(\ell)}$ is the input feature of $\ell$-th layer, $\mathbf{W}_k$ the learable filter, and $\mathbf{D}_k$ the degree matrix for hop $k$. 
The attention map \(\mathcal{M}^{(\ell)}\) is dynamically generated via ConvLSTM:
\begin{align}
\mathbf{C}_t^{(\ell)} &= \text{ConvLSTM}\left(\mathbf{F}^{(\ell)}_{t-1}, \mathbf{F}^{(\ell)}_t\right) \\
\mathcal{M}^{(\ell)} &= \sigma\left(\mathbf{C}_t^{(\ell)} \mathbf{W}_a^{(\ell)} + \mathbf{b}_a^{(\ell)}\right)
\end{align}
where \(\sigma\) is the sigmoid activation, \(\mathbf{W}_a^{(\ell)}, \mathbf{b}_a^{(\ell)}\) are learnable parameters, and \(\mathbf{F}^{(\ell)}_t\) denotes feature maps at frame \(t\).
This enhancement allows the network to focus on biomechanically critical areas and improves interpretability by visualizing $\mathcal{M}^{(\ell)}$ overlays on the skeleton.

\subsubsection{Biomechanical Features Stream}
The biomechanical features stream processes the sequence of 45-dimensional biomechanical features $\mathbf{B}$, derived from the raw pose data as described in previous section. This stream employs a two-layer Bidirectional LSTM (BiLSTM) with 128 hidden units in each direction to capture temporal dependencies from both past and future contexts in the sequence.

The BiLSTM processes the entire sequence of 90 frames, and the output from the last time step is passed through a fully-connected layer with ReLU activation. This layer maps the BiLSTM output to an 80-dimensional feature vector $\mathbf{H}_{\text{Biomech}} \in \mathbb{R}^{1 \times 80}$, aligning with the ST-GCN stream's output for the subsequent fusion mechanism.
\begin{equation}
    \mathbf{H}_{\text{Biomech}} = \text{FC}(\text{BiLSTM}(\textbf{B}))
\end{equation}
where $\mathbf{B} \in \mathbb{R}^{90 \times 45}$ is the input biomechanical feature sequence, and FC denotes the fully-connected layer.
These two streams are complementary: the ST-GCN captures spatial and temporal patterns directly from the pose data, while the BiLSTM focuses on temporal dynamics in the engineered biomechanical features, providing a richer representation for fall prediction.

\subsubsection{Feature Fusion and Prediction}

To effectively integrate the information from the ST-GCN and biomechanical feature streams, we introduce the \textbf{AttFusion Module}, a cross-attention mechanism designed to produce a unified representation for the final prediction task. This module leverages the rich spatio-temporal patterns from the ST-GCN stream to dynamically query and re-weight the most salient biomechanical features.


As illustrated in the Figure~\ref{fig:model_pipeline}, the AttFusion Module employs a dot-product attention mechanism. The feature vector from the ST-GCN stream, $\mathbf{H}_{\text{ST-GCN}}$, serves as the Query \textbf{(Q)}, while the feature vector derived from the biomechanical data, $\mathbf{H}_{\text{Biomech}}$, provides both the Key \textbf{(K)} and the Value \textbf{(V)}. Mathematically, these are formulated as $\mathbf{Q} = W_q \mathbf{H}_{\text{ST-GCN}}$, $\mathbf{K} = W_k \mathbf{H}_{\text{Biomech}}$, and $\mathbf{V} = W_v \mathbf{H}_{\text{Biomech}}$, where $W_q$, $W_k$, and $W_v$ are learnable weight matrices. The model learns to match the query against the keys to determine how much attention to pay to each part of the value vector.

The core attention operation is formulated as:
\begin{align}
\mathbf{H}^{hd}_{\text{Att}} = \text{Attention}(\mathbf{Q}, \mathbf{K}, \mathbf{V}) = \text{softmax}\left(\frac{\mathbf{Q}\mathbf{K}^T}{\sqrt{d_k}}\right)\mathbf{V}
\end{align}
where $d_k$ is the dimension of the key vectors, and $\mathbf{H}^{hd}_{\text{Att}}$ is the resulting attention-weighted feature vector for each head---a refined version of the biomechanical features, conditioned on the skeletal spatio-temporal context.

Following the attention calculation, this attended feature vector for all heads, $\mathbf{H}^{hd}_{\text{Att}}$, is combined and then passed through a linear transformation to create a preliminary fused representation:
\begin{align}
\mathbf{H}_{\text{Fused}}' = W_f \left[ \mathbf{H}^{\text{1}}_{\text{Att}} \parallel \mathbf{H}^{\text{2}}_{\text{Att}} \dots \parallel \mathbf{H}^{\text{hd}}_{\text{Att}} \right] + b_f
\end{align}
where $W_f$ and $b_f$ are the weights and bias of the linear layer, $\text{hd}$ is the total number of heads, and $\parallel$ denotes concatenation.

Finally, to ensure stable training and preserve the original ST-GCN information, this representation is combined with the initial ST-GCN feature vector via a residual connection to produce the final fused output, $\mathbf{H}_{\text{Fused}}$. This vector is then used for the final fall classification.

The resulting feature vector is then processed by a fully connected (FC) network to compute the probability of a fall $\mathbf{\hat{y}}$:
\begin{equation}
    \mathbf{\hat{y}} = \sigma(\mathbf{W}_{\text{FC}} \mathbf{H}_{\text{Fused}} + \mathbf{b}_{\text{FC}})
\end{equation}
where $\mathbf{W}_{\text{FC}} \in \mathbb{R}^{1 \times 160}$ is the weight matrix, and $\mathbf{b}_{\text{FC}}$ is the bias term. This prediction layer is structured as a multi-layer perceptron (256-128-64-1 units) with ReLU activations and dropout ($0.5$) applied between layers, except for the final output layer which uses sigmoid.

\subsection{Interpretability via Attention Mechanisms}
To understand model decision-making, we interpret the dynamic attention maps $\mathcal{M}^{(\ell)}$ from the Body Attention mechanism in the ST-GCN stream. Specifically, we extract attention weights $\alpha_{v,t}$ from the second Body Attention block, where spatial-temporal representations are most mature. Attention for each joint $v$ at frame $t$ is calculated as:
\begin{equation}
\alpha^{(\ell)}_{v,t} = \frac{\exp(f(\mathbf{F}^{(\ell)}_{v,t}))}{\sum_{v' \in \mathcal{V}} 
\exp(f(\mathbf{F}^{(\ell)}_{v',t}))}
\end{equation}
where $f(\cdot)$ is a learned scoring function, $\mathbf{F}^{(\ell)}_{v,t}$ 
is the feature vector for joint $v$ at frame $t$ in layer $\ell$, and 
$\mathcal{V} = \{1, 2, \ldots, 33\}$ is the set of all joints.
For effective visualization and comparison, these raw attention weights are normalized across both spatial (joints) and temporal (frames) dimensions using min-max normalization, yielding comparable attention maps \cite{song2019}.
We interpret the learned attention patterns through: spatio-temporal attention matrices, attention-weighted skeleton overlays, and temporal attention profiles. To interpret these attention patterns, we establish a mapping between high-attention regions and established biomechanical instability indicators. Quantitatively, we employ attention entropy \cite{heo2018} to measure attention dispersion and attention faithfulness \cite{chefer2021} to evaluate the correlation between attention weights and prediction sensitivity. The results of this analysis are described in more detail in the result section (Figure \ref{fig:vis_attention}).

\section{Results}
\label{sec:results}
Our work utilizes two datasets. The first, the MCF-UA dataset, combines the University of Arizona (UA) Stunt dataset~\cite{alhammouri2025fall} with the Multiple Camera Fall (MCF) dataset~\cite{auvinet2010} for environmental and subject diversity. The second is the Multi Visual Modality Fall Detection (MUVIM) dataset, a public resource with 15 subjects performing 15 fall types and 20 activities of daily living, captured by four synchronized Azure Kinect DK cameras \cite{denkovski2022multi}. For all videos, we extracted 33 3D joint locations per frame using MediaPipe pose~\cite{bazarevsky2020blazepose}. Key dataset characteristics are summarized in Table \ref{tab:datasets_summary}.

\begin{table}[ht]
\centering
\caption{Summary of Datasets}
\label{tab:datasets_summary}
\resizebox{0.47\textwidth}{!}{%
\begin{tabular}{lcc}
\toprule
\textbf{Feature} & \textbf{MCF-UA Dataset} & \textbf{MUVIM Dataset} \\
\midrule
 \vspace{2pt}
Reference & \specialcellnormal{Al-Hammouri \textit{et al.} (2025)\\\& Auvinet \textit{et al.} (2010)} & Denkovski \textit{et al.} (2022) \\ \vspace{2pt}
Subjects & \specialcellnormal{4 stunt actors\\ 15 subjects} & Multiple subjects \\ \vspace{2pt}
Environment & \specialcellnormal{Assisted living\\ Controlled indoor lab}
& General indoor \\ \vspace{2pt}
Camera & Single 4K, 8 IP cameras & (RGB, Depth, Thermal) \\ \vspace{2pt}
Action Types & 16 (4 UA, 12 MCF) & 35 (15 falls, 20 ADLs) \\
\bottomrule
\end{tabular}%
}
\end{table}


To handle class imbalance, we used a weighted binary cross-entropy loss function while training the model. Model performance was evaluated using Precision, Recall, F1-Score, and Area Under the Precision-Recall Curve (AUPRC), metrics well-suited for imbalanced safety-critical tasks \cite{alhammouri2025fall,liu2025real}. We assessed performance at prediction horizons up to 4 seconds before impact and used three evaluation protocols: \textbf{\textit{(1)}} Baseline Comparison (standard 80/20 split), \textbf{\textit{(2)}} Zero-Shot Generalization (strict subject-disjoint split), and \textbf{\textit{(3)}} Few-Shot Personalization (fine-tuning on $K=1-5$ samples from a new subject).

\subsection{Baseline Models}
\label{subsubsec:baseline_models}
We compared our proposed BioST-GCN against several baselines. Feature-based models included a Support Vector Machine (SVM) \cite{cortes1995support} and Extreme Gradient Boosting (XGBoost) \cite{chen2016xgboost}. Deep learning baselines included a four-layer Neural Network (NN) \cite{rumelhart1986learning}, an LSTM network \cite{alhammouri2025fall}, and a standard Spatio-Temporal Graph Convolutional Network (ST-GCN) \cite{yan2018spatial}. Our BioST-GCN builds on the ST-GCN by integrating an attention-based fusion mechanism for biomechanical features.

\begin{table*}[ht]
\centering
\caption{Comparative performance analysis of fall prediction models on both MCF-UA and MUVIM datasets at 2s time offset}
\label{tab:model_benchmark}
\resizebox{\textwidth}{!}{%
\begin{tabular}{@{}lcccccccc@{}}
\toprule

& \multicolumn{4}{c}{MCF-UA} & \multicolumn{4}{c}{MUVIM} \\

\cmidrule(lr){2-5} \cmidrule(lr){6-9}

Model & \textbf{P(\%)} & \textbf{R(\%)} & \textbf{F1(\%)} & \textbf{AUPRC(\%)} & \textbf{P(\%)} & \textbf{R(\%)} & \textbf{F1(\%)} & \textbf{AUPRC(\%)} \\
\midrule

\multicolumn{9}{@{}l@{}}{\textit{Baseline Methods}} \\
\midrule

Neural Network \cite{rumelhart1986learning} & 35.7 & 13.2 & 19.2 & 29.0 & 30.1 & 15.5 & 20.5 & 25.5 \\
SVM \cite{cortes1995support}                 & 34.4 & 29.0 & 31.4 & 30.0 & 32.8 & 25.1 & 28.4 & 28.1 \\
XGBoost \cite{chen2016xgboost}              & 66.7 & 5.3 & 9.8 & 23.0 & 60.2 & 7.1 & 12.7 & 20.5 \\
LSTM \cite{alhammouri2025fall}                      & 78.3 & 76.5 & 77.4 & 78.6 & 70.5 & 72.3 & 71.4 & 75.1 \\
\midrule
\multicolumn{9}{@{}l@{}}{\textit{Ours}} \\
\midrule
ST-GCN                                    & 80.2 & 89.5 & 84.6 & 86.6 & 71.2 & 80.9 & 75.7 & 80.3 \\

\rowcolor{gray!10}
\textbf{BioST-GCN}                  & \textbf{84.4} & \textbf{94.3} & \textbf{89.1} & \textbf{91.1} & \textbf{72.9} & \textbf{83.6} & \textbf{77.9} & \textbf{82.3} \\
\bottomrule
\end{tabular}
}
\end{table*}

\subsection{Overall Performance Comparison}
As shown in Table \ref{tab:model_benchmark}, our BioST-GCN model achieved superior performance on the MCF-UA dataset, with an F1-score of 89.1\% and AUPRC of 91.1\%, representing a relative improvement of over 11\% on both metrics compared to the LSTM baseline \cite{alhammouri2025fall}. Traditional machine learning models like SVM and XGBoost performed poorly, with XGBoost showing extremely low recall (5.3\%), making it unsuitable. The vanilla ST-GCN significantly outperformed the LSTM, and our fusion-based BioST-GCN further improved performance, with the most notable gain in recall (94.3\%), which is critical for minimizing missed falls.


\begin{figure}[t]
    \centering
    \includegraphics[width=0.48\textwidth]{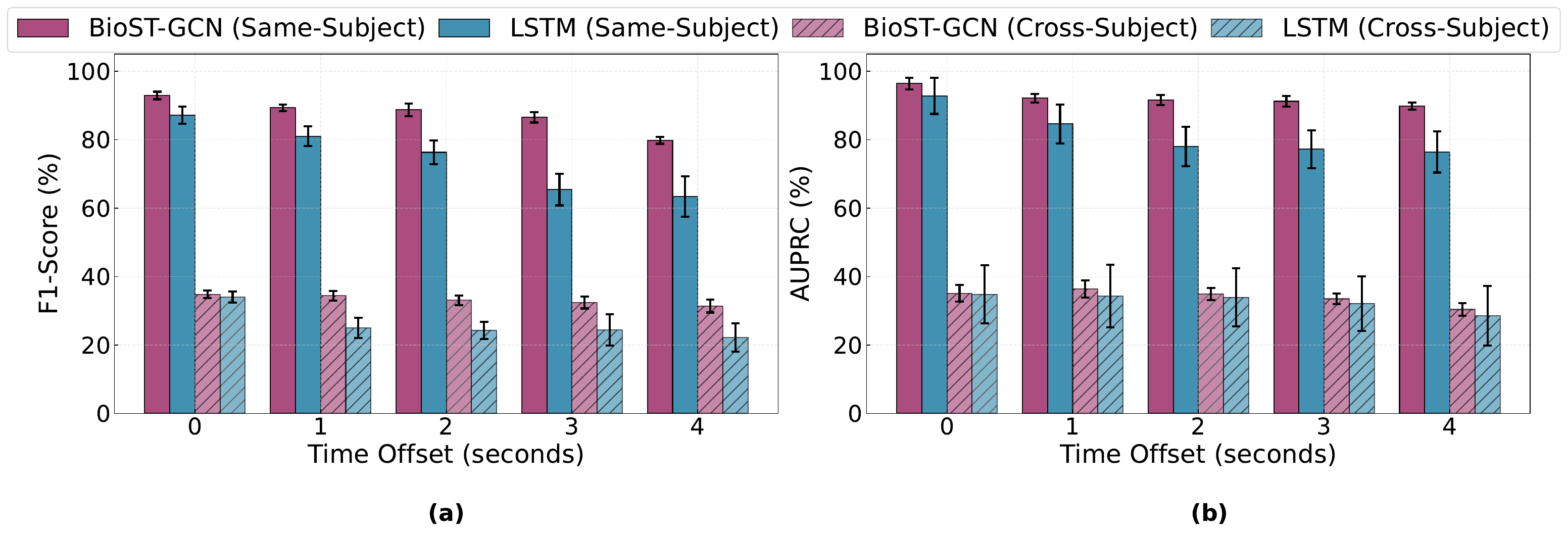}
    \caption{Performance comparison across prediction horizons (0-4 seconds before fall). 
    (a) F1-Score and (b) AUPRC for BioST-GCN and LSTM baseline in same-subject (split 1) and 
    cross-subject (split 2) settings. Error bars represent standard deviation over $5$ independent runs. 
    BioST-GCN demonstrates superior performance in same-subject settings and robust 
    generalization in cross-subject evaluation, with consistently lower variance than LSTM.}
\label{fig:results_split}
\end{figure}

\begin{figure*}[ht]
\centering
\includegraphics[width=\textwidth]{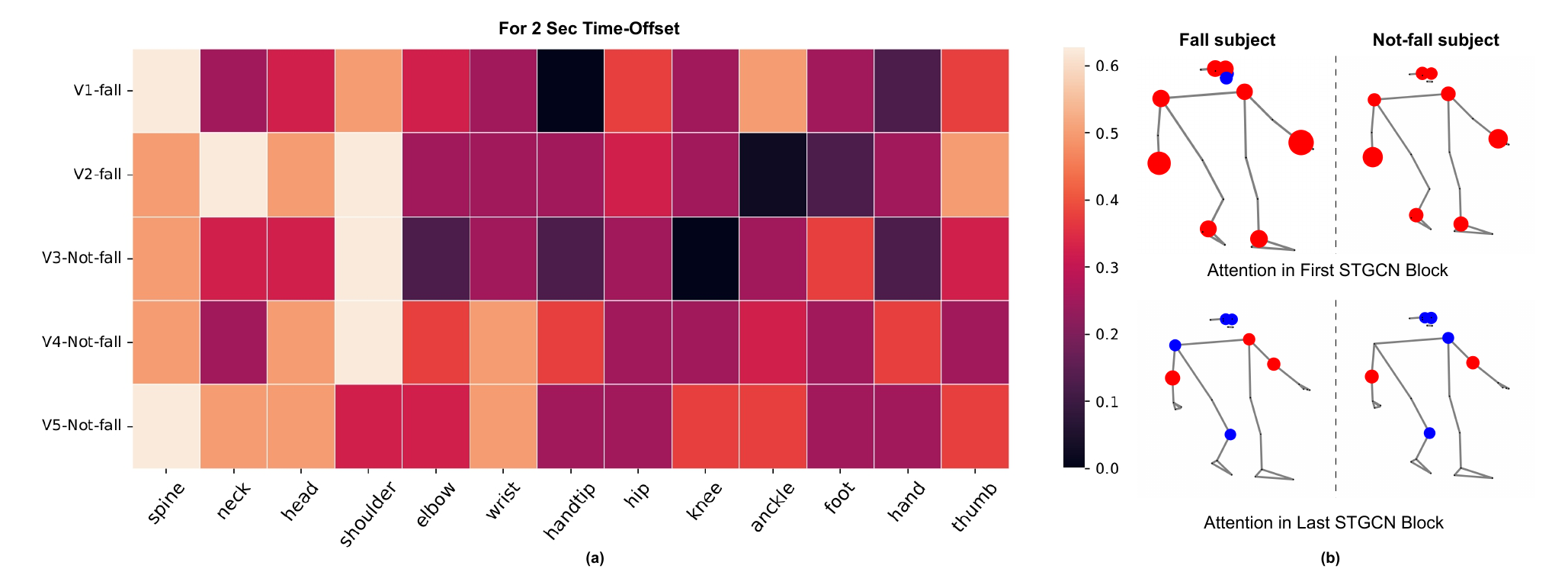}
\caption{Attention mechanisms in fall prediction. (a) Attention values (heat map) for different joints and scenarios (e.g., fall vs. non-fall), where color intensity represents attention strength. (b) Attention distribution across body joints in sequential ST-GCN blocks. The upper row depicts attention in an initial ST-GCN block, while the lower row shows attention in a deeper ST-GCN block. \textcolor{red}{Red circles} (attention values $> 0.3$) and \textcolor{blue}{blue circles} (attention values $< 0.3$) indicate the magnitude of attention received by individual joints, with circle size corresponding to attention magnitude.}
\label{fig:vis_attention}
\end{figure*}

\subsection{Performance Across Time-Offsets and Data Splits}

Figure~\ref{fig:results_split} compares BioST-GCN and LSTM performance under two split protocols. Split 1 (same-subject) measures accuracy on subjects observed during training; split 2 (cross-subject) assesses generalization to unseen individuals.
In the same-subject split, BioST-GCN achieves consistently high F1-Scores—$92.9 \pm 1.1\%$ at the fall moment ($t=0$s) and $86.5 \pm 1.5\%$ even at a 3-second horizon. In contrast, LSTM scores $87.1 \pm 2.5\%$ at $t=0$s, dropping to $65.4 \pm 4.6\%$ at 3 seconds. AUPRC values also reflect similar trends, with BioST-GCN reaching $96.4 \pm 1.7\%$ at $t=0$s and $91.2 \pm 1.6\%$ at 3 seconds, compared to the LSTM's $92.8 \pm 5.3\%$ and $77.2 \pm 5.5\%$ respectively.
In the cross-subject split, performance declines for both models. BioST-GCN records $34.8 \pm 1.1\%$ F1-Score at $t=0$s and $32.4 \pm 1.7\%$ at 3 seconds, while LSTM falls to $34.0 \pm 1.6\%$ and $24.4 \pm 4.6\%$. AUPRC values at $t=0$s are $35.1 \pm 2.4\%$ for BioST-GCN and $34.8 \pm 8.5\%$ for LSTM.
Thus, BioST-GCN consistently outperforms the baseline in same subjects and maintains greater robustness over longer prediction horizons. Nevertheless, both models shows considerable performance drops when evaluated on new subjects, highlighting the central challenge for generalizable fall prediction.


\subsection{Role of Joints in Fall Prediction}

Our model's improved performance is due to its ability to focus on biomechanically critical joints. Figure~\ref{fig:vis_attention}(a) shows that our model directs significantly more attention to trunk-related joints (mean: $0.71 \pm 0.08$) compared to less relevant distal extremities (mean: $0.23 \pm 0.15$). This aligns with established fall biomechanics, where displacement of the body's center of mass (COM), governed by the trunk and hips, is a primary fall indicator. The model further validates this by intensifying its focus on the trunk-hip relationship by $+27.3\%$ during correctly identified falls (true positives), confirming it learns meaningful anatomical cues. 
Furthermore, Figure~\ref{fig:vis_attention}(b) illustrates how this attention strategy evolves through the network's layers. While initial layers exhibit a broad attention distribution to capture overall motion, deeper layers progressively shift focus to the upper body. This hierarchical pattern is consistent across different subjects and prediction horizons, demonstrating that the model robustly learns to identify upper-body kinematics as the most salient features for imminent fall prediction.

\subsection{Ablation Studies}

\begin{table}[ht]
\centering
\caption{Component Analysis of the Proposed Model. We report F1-score (F1), and AUPRC, all in percentages (\%). The best results are highlighted in bold.}
\label{tab:component_analysis}
\begin{tabular}{RRRRRR}
\toprule
\specialcellbold{Is-\\Stacked?} & \specialcellbold{Aggr.\\Style} & \specialcellbold{Fusion\\Strategy} & \specialcellbold{Body\\Attention} & \textbf{F1} & \textbf{AUPRC} \\
\midrule
No  & Max Pool & Only Pose       & No & 82.6 & 84.1 \\
Yes & Max Pool & Only Pose       & No & 84.6 & 86.6 \\
Yes & Max Pool & Concat          & No & 85.4 & 87.5 \\
Yes & Max Pool & AttnFusion & No & 86.9 & 88.9 \\
Yes & LSTM     & AttnFusion & No & 88.1 & 90.2 \\
\textbf{Yes} & \textbf{BiLSTM}     & \textbf{AttnFusion} & \textbf{Yes} & \textbf{89.1} & \textbf{91.1} \\
\bottomrule
\end{tabular}%
\end{table}

\begin{figure}[ht]
  \centering
  \includegraphics[width=0.5\textwidth]{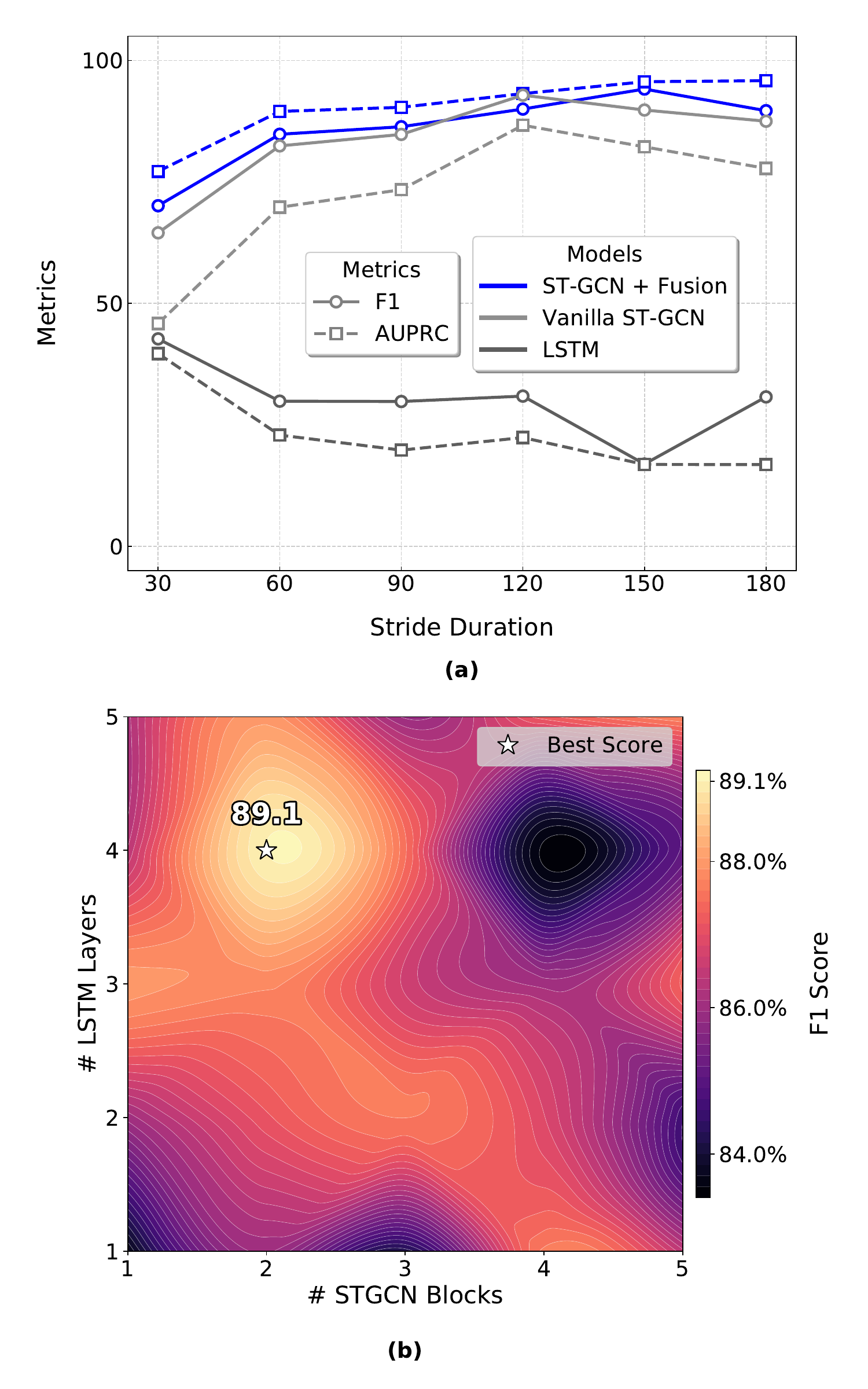}
  \caption{\textbf{(a)} Performance comparison with varying window sizes. \textbf{(b)} F1-Score performance for different numbers of STGCN blocks and LSTM layers.}
  \label{fig:exp_strides_hyper}
\end{figure}

\noindent\textbf{Component Analysis} In Table \ref{tab:component_analysis}, we report on different variations of our model to evaluate the contribution of each key component. We incrementally include or replace parts of the architecture, such as the use of a stacked layer of STGCN, the spatio-temporal feature aggregation style (Max Pool/LSTM), the feature fusion strategy, and a body-specific attention mechanism. First, we compare a plain (non-stacked) model versus a stacked model, both using Max Pool aggregation. We notice performance improves in the stacked case because of the network's ability to extract more complex and hierarchical features. On top of the stacked version, we then experiment with more advanced fusion strategies, moving from simple concatenation to our proposed AttentionFusion, which helps to better integrate the input modalities and improve performance. Results further improve when replacing the Max Pool aggregation with an LSTM, since it captures richer temporal information by capturing long-range dependencies. Finally, we include the Body Attention mechanism and BiLSTM, which is our final architecture. We achieve the best results with this setup because it allows the model to dynamically weigh the importance of different body regions while assessing an exercise, leading to a more robust and accurate evaluation.



\noindent\textbf{Changing Window Sizes} In Figure \ref{fig:exp_strides_hyper} (a), we analyze the impact of varying the temporal window size on model performance to determine the optimal setting. We observe that for both our proposed model, BioST-GCN and the Vanilla ST-GCN baseline, performance in terms of F1-Score and AUPRC generally improves as the window size increases from 40 to 120 frames. This trend suggests that shorter windows may not capture the complete temporal context of an action. However, performance begins to plateau and slightly decline for window sizes beyond 150 frames, likely because excessively long windows can introduce irrelevant frames or noise, diluting the discriminative features. Our proposed model consistently outperforms the baselines across all window sizes, and it achieves its peak performance within the $90-150$ frame range. Furthermore, with bigger window size, the model trends to take more time during inference. Based on this empirical analysis, we selected a window size of $90$ frames for our final model, as it provides the best balance between capturing sufficient temporal context and minimizing inference time constraint.

\noindent\textbf{Hyperparameter Analysis} We further conduct experiments for model specific hyper-parameters by varying the number of STGCN blocks and stacked LSTM layers in order to determine the optimal setups. In Figure \ref{fig:exp_strides_hyper} (b), we visualize the F1-Score across different model configurations. The results indicate that model performance is sensitive to the number of both STGCN blocks and LSTM layers. We observe that performance generally increases with the addition of layers, as deeper models can capture more complex spatio-temporal patterns. However, this trend does not continue indefinitely; an excessive number of layers leads to a decline in performance, likely due to overfitting or vanishing gradients. The highest F1-Score of 89.1\% is achieved with a configuration of 2 STGCN blocks and 4 LSTM layers. Therefore, we adopt this architecture for our final model to ensure the best performance.

\section{Conclusion}
\label{sec:conclusion}

This paper introduced BioST-GCN, a novel dual-stream architecture integrating Spatio-Temporal Graph Convolutional Networks with biomechanical feature processing for imminent fall prediction. On intra-subject evaluations (split 1), our model demonstrated strong performance, achieving an 89.1\% F1-score and a 91.1\% AUPRC, outperforming baseline approaches by over 11\%. Furthermore, the incorporated attention mechanism provided valuable interpretability, highlighting upper body kinematics.
A key finding of this work is the significant performance disparity when transitioning from intra-subject to cross-subject (split 2) evaluations. This underscores a critical challenge for the field: current models trained on simulated or limited datasets may learn subject-specific or simulation-specific cues rather than universally generalizable fall features. 
Although the BioST-GCN architecture represents a step forward, the significant gap in its performance across evaluation splits underscores a critical direction for future research. To develop robust, real-world fall prediction systems, future work must advance in two main areas. First, we need to develop efficient methods for personalizing models to an individual's movement patterns using minimal data. Second, it is equally crucial to improve the underlying data by acquiring more diverse examples and developing techniques to mitigate biases in existing datasets.
Addressing these fundamental data and generalization challenges, alongside continued algorithmic innovation, will be essential for translating vision-based fall prediction into effective clinical tools.

\appendix

\section{Appendix}

\subsection{Biomechanical Feature Extraction}
\label{sec:biomechanical_features}
\begin{figure*}[t]
    \centering
    \includegraphics[width=0.9\textwidth]{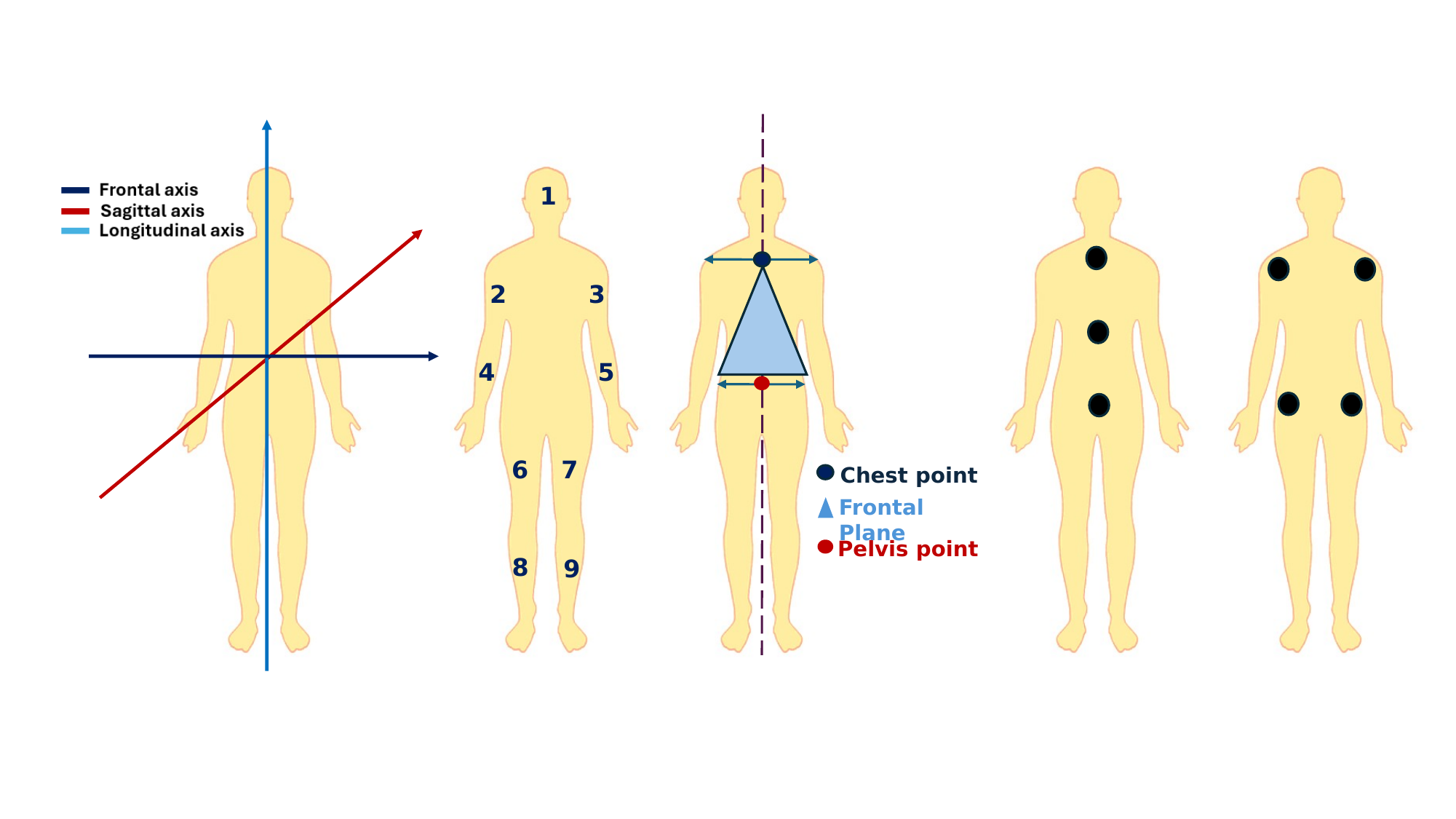}
    \caption{Illustration of the anatomical reference frame and body landmarks. The principal axes (frontal, sagittal, longitudinal) are shown on the left. The frontal plane is defined by chest and pelvis points.}
    \label{fig:anatomical_planes}
\end{figure*}

From the 3D skeletal landmarks, we derive a 45-dimensional biomechanical feature vector for each time frame. These features provide a meaningful representation of the subject's posture and movement dynamics, organized into four distinct categories as detailed below.




\textbf{Anatomical Reference}

The biomechanical analysis is based on a standard anatomical reference frame, as illustrated in Figure~\ref{fig:anatomical_planes}. The key assumptions for this frame are:

\begin{itemize}
    \item \textbf{Vertical axis (Y):} This axis is aligned with the vector of gravity.
    \item \textbf{Frontal plane:} This plane is defined by the positions of landmarks on the chest and pelvis, as shown in Figure~\ref{fig:anatomical_planes}.
    \item \textbf{Sagittal plane:} This plane is defined as being perpendicular to the frontal plane.
    \item \textbf{Upper/Lower Body Separation:} For the purpose of centroid calculation, the body is segmented into upper and lower parts at the level of the waist.
\end{itemize}

\textbf{Angular Positions (27 features)}

These features describe the orientation of nine key body segments in three rotational degrees of freedom: flexion/extension, abduction/adduction, and internal/external rotation. The calculation requires defining a local coordinate system for the proximal joint of each segment relative to its parent segment. For a given segment vector $\vec{v}_{\text{seg}}$ originating from a proximal joint with an established coordinate system ($\hat{\mathbf{x}}$: medial-lateral, $\hat{\mathbf{y}}$: longitudinal, $\hat{\mathbf{z}}$: anterior-posterior), the angles are defined as:

\begin{itemize}
    \item \textbf{Flexion/Extension ($\theta_{\text{flex/ext}}$):} Rotation around the medial-lateral ($\hat{\mathbf{x}}$) axis.
    \begin{equation}
        \theta_{\text{flex/ext}} = \arctan2(-\vec{v}_{\text{seg}} \cdot \hat{\mathbf{z}}, \vec{v}_{\text{seg}} \cdot \hat{\mathbf{y}})
    \end{equation}
    \item \textbf{Abduction/Adduction ($\theta_{\text{abd/add}}$):} Rotation around the anterior-posterior ($\hat{\mathbf{z}}$) axis.
    \begin{equation}
        \theta_{\text{abd/add}} = \arctan2(\vec{v}_{\text{seg}} \cdot \hat{\mathbf{x}}, \vec{v}_{\text{seg}} \cdot \hat{\mathbf{y}})
    \end{equation}
    \item \textbf{Internal/External Rotation ($\theta_{\text{rot}}$):} Rotation around the segment's longitudinal axis.
\end{itemize}

This set of three angles is computed for the following nine body segments, yielding $9 \times 3 = 27$ features:
\begin{enumerate}
    \item Trunk: $(\theta_{\text{trunk}}^{\text{flex/ext}}, \theta_{\text{trunk}}^{\text{abd/add}}, \theta_{\text{trunk}}^{\text{rot}})$
    \item Left \& Right Upper Arm: $(\theta_{\text{upper\_arm}}^{\text{flex/ext}}, \theta_{\text{upper\_arm}}^{\text{abd/add}}, \theta_{\text{upper\_arm}}^{\text{rot}})$
    \item Left \& Right Forearm: $(\theta_{\text{forearm}}^{\text{flex/ext}}, \theta_{\text{forearm}}^{\text{abd/add}}, \theta_{\text{forearm}}^{\text{rot}})$
    \item Left \& Right Thigh: $(\theta_{\text{thigh}}^{\text{flex/ext}}, \theta_{\text{thigh}}^{\text{abd/add}}, \theta_{\text{thigh}}^{\text{rot}})$
    \item Left \& Right Shank: $(\theta_{\text{shank}}^{\text{flex/ext}}, \theta_{\text{shank}}^{\text{abd/add}}, \theta_{\text{shank}}^{\text{rot}})$
\end{enumerate}

\textbf{Centroid Locations (9 features)}

These features are the 3D coordinates of the Center of Mass (COM) for the upper, lower, and total body. They are calculated as a weighted average of individual segment COMs ($C_i$), with masses ($m_i$) derived from Plagenhoef's anthropometric data \cite{plagenhoef1983}. The general formula is:
\begin{equation}
    C = \frac{\sum_{i} m_i C_i}{\sum_{i} m_i}
\end{equation}
The segment masses are defined as fractions of the total body mass ($M_{\text{total}}$).

The three centroids are computed as follows:
\begin{enumerate}
    \item \textbf{Upper Body Centroid ($C_{\text{upper}}$):} The COM of the head, trunk, upper arms, forearms, and hands, resulting in a 3D vector $(x_{\text{upper}}, y_{\text{upper}}, z_{\text{upper}})$.
    \item \textbf{Lower Body Centroid ($C_{\text{lower}}$):} The COM of the thighs, shanks, and feet, resulting in a 3D vector $(x_{\text{lower}}, y_{\text{lower}}, z_{\text{lower}})$.
    \item \textbf{Total Body Centroid ($C_{\text{total}}$):} The COM of all body segments, resulting in a 3D vector $(x_{\text{total}}, y_{\text{total}}, z_{\text{total}})$.
\end{enumerate}

\textbf{Yaw Trunk Angle (1 feature)}

This feature measures trunk rotation in the horizontal (transverse) plane. It is calculated from the 3D coordinates of the left ($P_{\text{L\_shoulder}}$) and right ($P_{\text{R\_shoulder}}$) shoulder landmarks projected onto the XZ plane (assuming Y is the vertical axis).
\begin{equation}
    \theta_{\text{yaw}} = \arctan2(p_{z,R} - p_{z,L}, p_{x,R} - p_{x,L})
\end{equation}
The $\arctan2$ function provides an unambiguous angle over a $360^{\circ}$ range.

\textbf{2D Image Coordinates (8 features)}

To retain perspective and elevation cues from the original image plane, the 2D pixel coordinates of four key landmarks are included as features. These eight values are:
\begin{enumerate}
    \item Left Hip: $(x_{\text{L\_hip}}, y_{\text{L\_hip}})$
    \item Right Hip: $(x_{\text{R\_hip}}, y_{\text{R\_hip}})$
    \item Left Shoulder: $(x_{\text{L\_shoulder}}, y_{\text{L\_shoulder}})$
    \item Right Shoulder: $(x_{\text{R\_shoulder}}, y_{\text{R\_shoulder}})$
\end{enumerate}

\subsection{Few-shot Analysis on Human Fall Prediction Datasets}
\label{sec:few_shot_analysis}

To evaluate the adaptability of our model to new, unseen individuals, we conducted a few-shot personalization experiment. This scenario is crucial for real-world applications, where a generalized model's performance can degrade on subjects with unique movement patterns. For this analysis, we fine-tuned our top-performing \textbf{BioST-GCN} model on a small number of fall samples ($K$) from subjects not included in the original training set.

The results, presented in Table \ref{tab:few_shot}, highlight the impact of personalization. The average ``Zero-Shot" ($K=0$) performance on new subjects is modest, with an F1-score of 33.31\%, indicating that the generalized model is not robust enough for direct deployment. However, the model's performance improves dramatically with minimal fine-tuning. By introducing just a single personalized sample ($K=1$), the F1-score has huge impact score 53.2\%, a relative improvement of nearly 60\%. This trend continues as more samples are added, with the F1-score reaching 82.5\% with only five samples. This demonstrates that our model, while requiring personalization, can be rapidly and efficiently adapted to achieve high performance for new users, turning a modest baseline into a robust and practical fall prediction system.

\begin{table}[ht]
\centering
\caption{Few-shot personalization performance of BioST-GCN on new subjects from the MCF-UA dataset. $K$ denotes the number of fall samples used for fine-tuning. $K=0$ represents the average zero-shot performance.}
\label{tab:few_shot}
\resizebox{0.45\textwidth}{!}{%
\begin{tabular}{@{}lcccc@{}}
\toprule
\textbf{Fine-tuning Samples ($K$)} & \textbf{P(\%)} & \textbf{R(\%)} & \textbf{F1(\%)} & \textbf{AUPRC(\%)} \\
\midrule
$K=0$ (Zero-Shot) & 28.6 & 41.2 & 33.3 & 30.3 \\
$K=1$ & 45.0 & 65.0 & 53.2 & 55.5 \\
$K=2$ & 60.1 & 75.3 & 66.8 & 68.2 \\
$K=3$ & 70.5 & 82.1 & 75.9 & 78.3 \\
$K=4$ & 75.2 & 86.4 & 80.4 & 82.6 \\
$K=5$ & \textbf{78.1} & \textbf{87.5} & \textbf{82.5} & \textbf{85.1} \\
\bottomrule
\end{tabular}%
}
\end{table}

\subsection{Fall Prediction Performance Across Time Horizons}
Figure~\ref{fig:results_split} in main paper presents the F1-Score and AUPRC of BioST-GCN compared to baseline models across prediction horizons from 0s to 4s in both same-subject and cross-subject settings. To establish statistical rigor, all experiments were repeated 5 times with different random seeds; detailed results with mean $\pm$ standard deviation are provided in Tables~\ref{tab:f1_all_horizons_with_drop} and~\ref{tab:auprc_all_horizons_with_drop}.

The low variance across runs (typically $<2\%$ for BioST-GCN) confirms the stability and reliability of our model. In the same-subject setting, BioST-GCN achieves an F1-Score of \(92.9 \pm 1.1\%\) at 0s and \(79.8 \pm 1.0\%\) at 4s, substantially outperforming the LSTM baseline. In the more challenging cross-subject setting, BioST-GCN maintains consistent performance (\(34.8 \pm 1.1\%\) at 0s, \(31.4 \pm 1.9\%\) at 4s), demonstrating robust generalization to unseen subjects.

\begin{table*}[ht]
\centering
\caption{F1-Score (\%, mean $\pm$ std. dev. over 5 runs) and Relative Drop at all prediction horizons.}
\label{tab:f1_all_horizons_with_drop}
\begin{tabular}{@{}llccccc@{}}
\toprule
\textbf{Setting} & \textbf{Model} & \textbf{0s} & \textbf{1s} & \textbf{2s} & \textbf{3s} & \textbf{4s} \\ \midrule
\addlinespace[0.3em]
Same-Subject & LSTM & 87.1 $\pm$ 2.5 & 81.0 $\pm$ 2.9 & 76.3 $\pm$ 3.5 & 65.4 $\pm$ 4.6 & 63.4 $\pm$ 5.9 \\
& BioST-GCN &  \textbf{92.9 $\pm$ 1.1} & \textbf{89.3 $\pm$ 1.0} & \textbf{88.7 $\pm$ 1.8} & \textbf{86.5 $\pm$ 1.5} & \textbf{79.8 $\pm$ 1.0} \\
\addlinespace[0.6em]
Cross-Subject & LSTM & 34.0 $\pm$ 1.6 & 25.0 $\pm$ 2.9 & 24.3 $\pm$ 2.5 & 24.4 $\pm$ 4.6 & 22.2 $\pm$ 4.1 \\
& BioST-GCN & \textbf{34.8 $\pm$ 1.1} & \textbf{34.4 $\pm$ 1.4} & \textbf{33.1 $\pm$ 1.4} & \textbf{32.4 $\pm$ 1.7} & \textbf{31.4 $\pm$ 1.9} \\
\addlinespace[0.6em] \midrule \addlinespace[0.3em]
\textbf{Relative Drop (\%)} & LSTM & 61.0 & 69.1 & 68.2 & 62.7 & 65.0 \\
& BioST-GCN & 62.5 & 61.5 & 62.7 & 62.5 & 60.7 \\
\bottomrule
\end{tabular}
\end{table*}

\begin{table*}[ht]
\centering
\caption{AUPRC (\%, mean $\pm$ std. dev. over 5 runs) and Relative Drop at all prediction horizons.}
\label{tab:auprc_all_horizons_with_drop}
\begin{tabular}{@{}llccccc@{}}
\toprule
\textbf{Setting} & \textbf{Model} & \textbf{0s} & \textbf{1s} & \textbf{2s} & \textbf{3s} & \textbf{4s} \\ \midrule
\addlinespace[0.3em]
Same-Subject & LSTM & 92.8 $\pm$ 5.3 & 84.6 $\pm$ 5.7 & 78.0 $\pm$ 5.8 & 77.2 $\pm$ 5.5 & 76.4 $\pm$ 6.0 \\
& BioST-GCN & \textbf{96.4 $\pm$ 1.7} & \textbf{92.1 $\pm$ 1.3} & \textbf{91.6 $\pm$ 1.5} & \textbf{91.2 $\pm$ 1.6} & \textbf{89.8 $\pm$ 1.0}  \\
\addlinespace[0.6em]
Cross-Subject & LSTM & 34.8 $\pm$ 8.5 & 34.3 $\pm$ 9.1 & 33.9 $\pm$ 8.5 & 32.1 $\pm$ 8.0 & 28.5 $\pm$ 8.7 \\
& BioST-GCN & \textbf{35.1 $\pm$ 2.4} & \textbf{36.4 $\pm$ 2.5} & \textbf{34.9 $\pm$ 1.8} & \textbf{33.5 $\pm$ 1.6} & \textbf{30.4 $\pm$ 1.8} \\
\addlinespace[0.6em] \midrule \addlinespace[0.3em]
\textbf{Relative Drop (\%)} & LSTM & 62.5 & 59.5 & 56.5 & 58.4 & 62.7 \\
& BioST-GCN & 63.6 & 60.5 & 61.9 & 63.3 & 66.1 \\
\bottomrule
\end{tabular} 
\end{table*}

\subsection{Generalization Analysis: Impact of Biomechanical Features}
\label{sec:generalization_analysis}

To investigate whether the engineered 45-dimensional biomechanical features increase subject-specificity and harm generalization (as hypothesized by reviewers), we calculated the relative performance drop from the same-subject to cross-subject setting for both BioST-GCN and the LSTM baseline. This metric quantifies how much performance degrades when the model encounters unseen subjects.

As shown in Figure~\ref{fig:results_split} in main paper, despite BioST-GCN starting from a significantly higher performance baseline in the same-subject setting (e.g., 88.7\% F1-Score vs. LSTM's 76.3\% at 2s), its relative performance drop is remarkably consistent with that of the simpler LSTM model across all time offsets. For F1-Score, BioST-GCN's relative drop ranges from 60.7\% to 62.7\% (mean: 61.9\%), while LSTM's ranges from 61.0\% to 69.1\% (mean: 65.2\%). Similar patterns are observed for AUPRC.

This evidence contradicts the hypothesis that detailed biomechanical features exacerbate subject-specificity. Instead, our feature set provides a robust foundation that generalizes as effectively as simpler models while enabling significantly higher absolute performance when personalization is possible. The Body Attention mechanism and dual-stream architecture contribute to this generalization capability by learning anisotropic, context-aware joint representations rather than overfitting to subject-specific patterns.

\subsection{Component-Level Ablation: Analysis of 0s Offset Performance}
\label{sec:ablation_0s}

At the 0s cross-subject offset, LSTM and BioST-GCN achieve statistically equivalent F1-scores (34.0\% vs 34.8\%) despite BioST-GCN's advantages at longer horizons. To understand this equivalence, we conducted a five-run ablation study of each component:

\begin{table}[ht]
\centering
\caption{Component-level ablation at 0s offset (Cross-Subject).}
\label{tab:ablation_0s}
\begin{tabular}{@{}lc@{}}
\toprule
\textbf{Model Component} & \textbf{F1-Score (\%)} \\ \midrule
BiLSTM only & 34.0 
$\pm$ 2.6 \\
ST-GCN only & 34.6 
$\pm$ 1.1 \\
BioST-GCN (AttFusion) & 34.8 
$\pm$ 1.1 \\
\bottomrule
\end{tabular}
\end{table}

The key finding is not performance differences, but prediction stability: BiLSTM exhibits twofold higher variance (std = 2.6\%) compared to ST-GCN approaches (std = 1.1\%). AttFusion leverages ST-GCN's reliable signals while maintaining performance. This performance plateau is specific to zero-shot detection; at 
$t>0$, ST-GCN's spatio-temporal structure enables BioST-GCN to substantially outperform baselines (+9.1\% at 1s, +9.0\% at 4s).

\subsection{Discussion}
\textbf{The Personalization Imperative and the Path to Generalizable Fall Prediction}
\label{subsec:personalization_generalization}

Our study clearly demonstrates a significant generalization gap between intra-subject (split 1) and cross-subject (split 2) testing scenarios. While models trained and evaluated on data incorporating the same individuals (split 1) achieved a high F1-score of 89.05\%, this performance substantially declined to 35.90\% when models were tested on entirely new subjects (split 2). This pronounced difference does not solely reflect limitations of a specific architecture; rather, it underscores a fundamental challenge in applying machine learning to highly variable human movement: models may learn features highly specific to the individuals or the nature of the data present in the training set, rather than capturing universally generalizable fall precursor patterns.
Individual variations in movement strategies, biomechanics, and responses to instability are considerable. For real-world applicability, particularly in clinical settings, achieving robust fall prediction necessitates strategies that can account for this heterogeneity. Our findings strongly suggest two convergent paths forward: first, the development of efficient personalization techniques capable of adapting a general model to an individual's unique movement signature with minimal new data; and second, a concerted effort to address inherent biases in currently available training data.

A primary contributor to the observed generalization challenges, especially when the ultimate goal is deployment for vulnerable elderly populations, is the field's common reliance on simulated fall data. Ethical and logistical hurdles in collecting sufficient real-world fall data from target groups (e.g., frail older adults, individuals with cognitive impairments or conditions such as diabetes) have led to the pragmatic use of younger, healthy 'stunt actors' in controlled laboratory settings. While facilitating safe data acquisition, this practice, as our results indirectly highlight, may lead models to learn patterns fundamentally different from those underlying real-world, unexpected falls in the target demographic.
This discrepancy gives rise to a critical, testable hypothesis regarding `Stunt Actor Bias': that the kinematic and biomechanical features most predictive of simulated falls in young actors may not directly translate, or may manifest differently, in genuine falls experienced by older adults with varying health conditions \cite{manor2010physiological}. The profound physiological and functional differences between these groups—spanning cognitive processing during unexpected events \cite{yogev2008role, riederer2017diabetic, eng2008balance, johnston2012association, whitmer2009hypoglycemic, maidan2018age}, musculoskeletal and neuromuscular capacities \cite{eng2008balance, oddsson2020effects}, frailty and physiological reserves \cite{dunning2020care, soysal2024prevalence}, and intrinsic gait dynamics and fall kinematics \cite{james2016gait, jason2022effects, wolfe2022effects}—form the basis of this hypothesis. For instance, the intentional, controlled nature of a simulated fall by a robust actor likely lacks the cascading physiological failures (e.g., sensory loss, muscle weakness, impaired central processing) and environmental interactions that characterize many real-world falls in frail individuals.

Our work, by demonstrating strong performance with the proposed BioST-GCN model on simulated data (split 1) and then quantifying the performance drop on unseen subjects (split 2), provides a valuable baseline and a methodological framework. It sets the stage for future research to explicitly test this Stunt Actor Bias. Such research could investigate whether the key predictive features identified by our model (e.g., trunk-hip dynamics via attention) remain equally salient, or require realignment, when applied to (eventually available) real-world fall data from diverse elderly cohorts. Successfully identifying and mitigating this bias, perhaps through domain adaptation, transfer learning from more representative (even if smaller) datasets, or advanced simulation refinement, is paramount for translating the promise of vision-based fall prediction into reliable clinical tools. Our findings highlight the urgency and provide a data-driven impetus for these next critical steps in research.
\bibliography{aaai2026}

\end{document}